\documentclass[11pt]{article}

\usepackage{EMNLP2022}

\usepackage{times}
\usepackage{latexsym}

\usepackage[T1]{fontenc}

\usepackage[utf8]{inputenc}

\usepackage{microtype}

\usepackage{inconsolata}

\usepackage{amsmath}
\usepackage{amssymb}
\usepackage{graphicx}
\usepackage{multirow}
\usepackage{booktabs}

\newif\ifcomments
\commentstrue
\ifcomments
\usepackage[normalem]{ulem}
\definecolor{CMpurple}{rgb}{0.6,0.18,0.64}
\newcommand\cm[1]{\textcolor{CMpurple}{\textsf{\scriptsize[\textbf{CM\@:} #1]}}}
\newcommand\cmi[1]{\textcolor{CMpurple}{#1}}
\newcommand\cmm[1]{\marginpar{\raggedright\tiny\textcolor{CMpurple}{\textsf{{\bfseries CM\@:} #1}}}}
\newcommand\cms{\bgroup\markoverwith{\textcolor{CMpurple}{\rule[.4ex]{2pt}{0.8pt}}}\ULon}
\else
\newcommand\cm[1]{}
\newcommand\cmi[1]{\ignorespaces}
\newcommand\cmm[1]{}
\newcommand\cms[1]{#1}
\fi

\title{Truncation Sampling as Language Model Desmoothing}

\newcommand{\AnD}{\hskip 2em plus 1fil minus 0.5em}
 \author{John Hewitt \AnD Christopher D. Manning \AnD Percy Liang \\
 Department of Computer Science\\
 Stanford University \\
 \{\texttt{johnhew,manning,pliang}\}\texttt{@cs.stanford.edu}}

\begin{document}
\maketitle

\begin{abstract}
Long samples of text from neural language models can be of poor quality.
Truncation sampling algorithms--like top-$p$  or top-$k$---address this by setting some words' probabilities to zero at each step.
This work provides framing for the aim of truncation, and an improved algorithm for that aim.
We propose thinking of a neural language model as a mixture of a true distribution and a \textit{smoothing} distribution that avoids infinite perplexity.
In this light, truncation algorithms aim to perform \textit{desmoothing}, estimating a subset of the support of the true distribution. Finding a good subset is crucial: 
 we show that top-$p$ unnecessarily truncates high-probability words, for example causing it to truncate all words but \textit{Trump} for a document that starts with \textit{Donald}.
We introduce $\eta$-sampling, which truncates words below an entropy-dependent probability threshold.
Compared to previous algorithms, $\eta$-sampling generates more plausible long English documents according to humans, is better at breaking out of repetition, and behaves more reasonably on a battery of test distributions. 
\end{abstract}

\section{Introduction}

The complex, long-range dependencies of natural language make its generation an outstanding challenge.
While there has been enormous progress on language modeling that has increased the coherence and length of generation \cite{brown2020language,chowdhery2022palm},
sampling directly from a language model can still result in nonsensical output \cite{Holtzman2020The,pillutla2021mauve}.

The most effective heuristics for generating high quality, diverse samples fall under a category we term \textit{truncation sampling}.
These algorithms set some words' probabilities to zero when generating each word \cite{fan-etal-2018-hierarchical,basu2021mirostat,meister-cotterell-2021-language}.
Methods differ by their truncation criteria, ranging from simple (keep the $k$ most likely) to complex, and all improve sample quality compared to direct sampling \cite{Holtzman2020The}.
We ask (1) what is the aim of truncation and (2) how can we improve it?

\begin{figure}
\centering
\includegraphics[width=\linewidth]{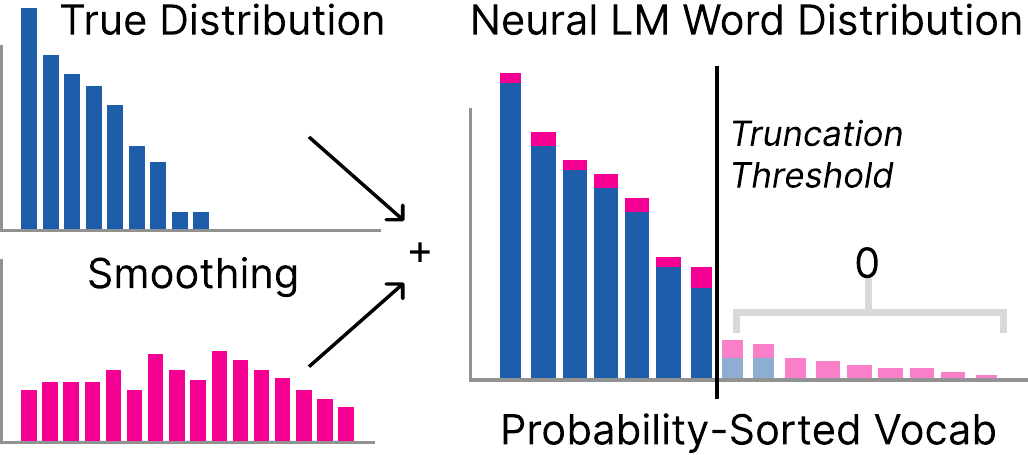}
\caption{\label{figure_fig1} A neural LM as a mixture of the true distribution, and a uniform-like smoothing distribution. Truncation aims to approximate the true distribution support.} %
\end{figure}

\begin{figure*}
  \centering
  \small
  \begin{tabular}{p{7.5cm}p{7.5cm}}
    \toprule
    \bf Unsmoothed 5-gram & \bf Smoothed 5-gram \\
    \midrule
    \ldots\ a quadcopter flight controller (RTFQ Flip MWC) that supports I2C sensors for adding thing like a barometer, magnetometer, and GPS system. The officially supported sensor block (BMP180, HMC5883L on one board) is discontinued, as far as I know, everyone involved lived to sing another day. &
    \ldots\ disorder and an extreme state of dysmetabolism characterized by extensive erythema and a significant reduction in uncovered {\color{red} Hawkingû McK 400 ruled restrainedcombeblow uncle cowork Carssoild Gareth focused <@ indecentlol by102 exchanged Volvo compositionsbackground prostate}\\
    \bottomrule
  \end{tabular}
\caption{\label{figure_ngram_samples} Portions of unconditional samples from an unsmoothed and uniform-smoothed $5$-gram model; divergence due to leaving the support of the high-order distribution is in red.}
\end{figure*}

Our key insight is to write a neural language model's distribution as a mixture of the true distribution and a uniform-like \textit{smoothing} distribution.
This idealized assumption is motivated by KL-divergence: models incur large KL at test time when they place near zero probability on an observed word \cite{kang-hashimoto-2020-improved}.
Through this lens, the goal of truncation is to \textit{desmooth}: to approximately recover the words on which the true distribution places some probability.

As a stark example of smoothing degenerating sample quality, we show that a $5$-gram language model smoothed with the uniform distribution generates nonsense as soon as a word is sampled from outside the support of the $5$-gram model (Figure~\ref{figure_ngram_samples}).
Intuitively, sampling outside the $5$-gram support causes future probabilities to be poorly estimated.

We derive principles of truncation from an explicit smoothing model that formalizes the intuition that (1) words with high probability should not be truncated, and (2) when all words in the distribution have low probability, only words with low probability relative to the rest should be truncated.
We find that state-of-the-art truncation sampling algorithms like top-$p$ break these principles.
For example, in top-$p$ truncation (e.g., $p=0.95$), the most likely few words can take up $p$\% of the distribution, causing the next-most likely word to be truncated even if it has high probability (e.g., $4$\%).

From our two truncation principles we derive $\eta$-sampling, a new algorithm that truncates any word whose probability under the LM is both (1) smaller than an absolute probability threshold and (2) smaller than a probability threshold that depends on the entropy of the distribution.
As we'll show, this ensures that, e.g., though GPT-2 large assigns probability $0.96$ to the word \textit{Trump} for a document starting with \textit{Donald}, $\eta$-sampling allows multiple possible continuations, unlike top-$p=0.95$.

We extensively study the behavior of $\eta$-sampling in comparison to top-$p$ sampling and typical decoding \cite{meister-cotterell-2021-language}.
Since each method allows for a range of quality-diversity tradeoffs, we set each method's hyperparameter by maximizing MAUVE score \cite{pillutla2021mauve}.
We find that $\eta$-sampling truncates more reasonably on a CheckList-style \cite{ribeiro-etal-2020-beyond} battery of distributions.
Top-$p$ and typical decoding over-truncate low-entropy distributions (like in the \textit{Donald} example).
Finally, $\eta$-sampling generates long documents that humans find more plausible and is better at breaking out of repetition.\footnote{Our code is available at \url{https://github.com/john-hewitt/truncation-sampling}.}

\section{Background}
\subsection{Language Models}
Let random variable $X=(X_1,\dots,X_T)$ denote a sequence of tokens, where each $X_i$ is in finite vocabulary $\mathcal{V}$.
We'll use $x_{<i}$ to refer to a specific prefix, $x_i$ a specific word in context, and $x$ an arbitrary word in $\mathcal{V}$.
An autoregressive language model (LM) is a distribution $P_\theta(X)$ indexed by parameters $\theta$ that is factorized as $P_\theta(x) = \prod_{i=1}^{T}P_\theta(x_i\mid x_{<i})$. 
We call $P_\theta(X_i \mid x_{<i})$ over $\mathcal{V}$ the conditional distribution of the LM given context $x_{<i}$.
An LM is trained to minimize the KL-divergence between (an empirical estimate of) the \textit{true} distribution $P^*(X)$ and $P_\theta(X)$.
Recent language models have achieved strikingly low (held-out) KL-divergence \cite{radford2019language}.

Language models are used not just to score the probability of existing sequences, but to generate sequences as $\hat{x} \sim P_\theta(X)$, a building block for tasks like summarization and long-form question answering \cite{fan2019eli5,liu-lapata-2019-text}.
However, to successfully generate high-variety, high-quality long samples from neural LMs on high-entropy distributions, it is currently necessary to reallocate probability from the tail of conditional distributions \cite{Holtzman2020The,pillutla2021mauve}.
Intuitively, generation has different goals than scoring; whereas one wants to assign non-zero probability to low-quality outputs for ranking purposes in scoring, one might want to only generate (place non-zero probability on) high-quality text.

\subsection{Truncation sampling}
There are many ways to reassign probability mass from the tail of the word-level distributions of a model to the head---like temperature scaling---but explicit truncation of low-probability words has been shown to be the most useful \cite{Holtzman2020The,pillutla2021mauve}.
Truncation sampling algorithms compute the following truncated distribution at each time step:
\par\nobreak
\vspace{-10pt}
{
  \setlength{\abovedisplayskip}{8pt}
  \setlength{\belowdisplayskip}{\abovedisplayskip}
  \setlength{\abovedisplayshortskip}{0pt}
  \setlength{\belowdisplayshortskip}{4pt}
\begin{align}
P_\text{trunc}(x\mid x_{<i}) = \begin{cases} P_\theta(x\mid x_{<i})/Z_{x_{<i}} & x \in \mathcal{A}_{x_{<i}}\\
0 & \text{o.w.}
\end{cases}
\end{align}
}%
where $\mathcal{A}_{x_{<i}}\subseteq \mathcal{V}$ we call the \textit{allowed set} for the algorithm for that prefix, and $Z_{x_{<i}}=\sum_{x\in\mathcal{A}_{x_{<i}}}P_\theta(x\mid x_{<i})$ is the renormalization term.%

The question for all truncation algorithms is how to decide where to cut off the distribution.
Top-$k$ sampling \cite{fan-etal-2018-hierarchical} keeps the $k$ most likely words.
Top-$p$ sampling \cite{Holtzman2020The} improved upon it by noting that sometimes more or fewer than $k$ words should be in the allowed set, instead allowing the minimal set of words to keep $p$ percent of the probability.
More recently, Mirostat adaptively truncates so as to achieve samples of a given probability \cite{basu2021mirostat}, and typical decoding truncates so as to locally match an informativeness criterion \cite{meister2022typical}.
We pursue an understanding of truncation as attempting to recover (a conservative estimate of) the \textit{true} training distribution $P^*$.

\section{Truncation as Desmoothing}

\subsection{KL-divergence and mode covering}
Language models are trained to minimize the KL-divergence to an empirical approximation of true distribution $P^*(X)$. 
Recall that the KL-divergence for a model's conditional distribution $P_\theta(X\mid x_{<i})$ to the true conditional distribution $P^*(X\mid x_{<i})$ is
\par\nobreak
{\small
  \setlength{\abovedisplayskip}{6pt}
  \setlength{\belowdisplayskip}{\abovedisplayskip}
  \setlength{\abovedisplayshortskip}{0pt}
  \setlength{\belowdisplayshortskip}{4pt}
\begin{align}
 \sum_{x\in \mathcal{V}} P^*(x\mid x_{<i}) \log \frac{P^*(x \mid x_{<i})}{P_\theta(x \mid x_{<i})} \label{kldiv}
\end{align}
}%
KL-divergence is known to be \textit{mode-covering}; it heavily penalizes errors of coverage.
When training from samples, an observed word $x_i$ in context $x_{<i}$ causes the model to incur a loss of $-\log P_\theta(x_i\mid x_{<i})$, which approaches infinity as the model probability approaches 0.\footnote{
Likewise during evaluation, the held-out perplexity $2^{\mathbb{E}_{x_i,x_{<i}}\log P_\theta(x_i\mid x_{<i})}$ is infinite if zero mass is placed on an observed word.}
Neural LMs use shared representations to generalize beyond the training data, e.g., knowing that the word \textit{home} may appear in a context where \textit{house} appeared.
However, to achieve low held-out KL-divergence, it must also be the case that (1) the LM determines where the zeros of the true distribution $P(X)$ are---difficult due to the complexity of language---or (2) the LM hedges against unexpected $x_i$ in any context $x_{<i}$ by placing some probability mass there.
Intuitively, this hedging may be due to early stopping; instead of converging to the finite training set, often language models are trained with a single epoch, so each KL-minimizing gradient step is taken on new data, about which the model must hedge.

\subsection{A neural LM as a smoothed distribution}
\label{section:smoothingmodel}
We present a framework for neural LMs wherein smoothing aids in KL-divergence minimization by placing a small amount of probability mass on all words.
Consider a true conditional distribution $P^*(X_i\mid x_{<i})$ over $\mathcal{V}$.
We think of the LM distribution $P_\theta(X_i\mid x_{<i})$ as the result of smoothing the true distribution with a distribution $Q(X_i\mid x_{<i})$ that is like the uniform distribution.
Specifically, we pose that the neural LM is a linear interpolation:
\begin{align}
P_\theta(X_i\mid x_{<i}) = & \lambda_{x_{<i}} P^*(X_i\mid x_{<i})  \nonumber \\
&+ (1-\lambda_{x_{<i}})Q(X_i\mid x_{<i}) \label{eqn_smoothing}
\end{align}
where $\lambda_{x_{<i}}\in (0,1]$ specifies the strength of the smoothing.
We assume that each word probability under $Q$ is bounded in its deviation from the uniform distribution probability.
For all $x\in\mathcal{V}$, we assume $Q(x\mid x_{<i})\in (\frac{1-\delta}{|\mathcal{V}|},\frac{1+\delta}{|\mathcal{V}|})$ where $\delta$ is a constant specifying non-uniformity.
We assume constraints on $\lambda_{x_{<i}}$ that reflect how the amount of smoothing should be (1) small and (2) dependent on how well-estimated a given conditional distribution is.
Specifically, we assume that $\lambda_{x_{<i}} \geq \max(\bar{\lambda}_{x_{<i}}, \bar{\lambda})$ where $\bar{\lambda}$ is a constant near 1 (e.g., $0.8$), independent of prefix.
The exact form we use for the context-dependent $\bar{\lambda}_{x_{<i}}$ is: $1-\frac{V\alpha \exp(-h_{x_{<i}})}{1+\delta}$.
As we will show later, this form implies that for a distribution of entropy $h$, words with probability $0$ under $P^*$ have probability bounded by $\alpha\exp(-h)$ under the language model.\footnote{Note that $\exp(-h)$ is the probability in a uniform distribution of entropy $h$. This entropy is of $P^*(X_i\mid x_{<i})$.}
A simple intuition for high-entropy distributions having less smoothing is that, e.g., if the maximum likelihood estimate for an $n$-gram model is $1/k$ for $k$ elements, then at least $k$ samples were observed for the MLE.\footnote{Even with this argument, the idea that high-entropy distributions are likely better estimated is probably the most tenuous assumption. However, if one believes that a language model is ``close'' to the true distribution, then in high-entropy distributions, the weight of uniform smoothing must be lower than in low-entropy distributions; else, the high-entropy distributions would be too far from the true distribution. Further, empirically, the highest-entropy distributions in language models, like \textit{A \ldots} or \textit{The \ldots} are high-entropy due to exceptional evidence (examples) of possible continuations. Put another way, this suggests the entropy is from epistemic uncertainty \cite{osband2022epistemic}.}

\subsection{A local measure of truncation quality}
\label{section:stv}
Under the smoothing model, we can make precise the tradeoff between (1) truncating too little, allowing words that are poor continuations, and (2) truncating too much and losing the diversity of the true distribution.
Let $ S_{x_{<i}}^* = \{x\in\mathcal{V} \mid P^*(x\mid x_{<i}) > 0\} $ be the true distribution support (set of words with non-zero probability) for the prefix $x_{<i}$.
Recall that $\mathcal{A}_{x_{<i}}\subseteq \mathcal{V}$ is the set of words allowed by a truncation algorithm, and that $P_\text{trunc}$ is the distribution of $P_\theta$ after truncation. %
Let $\overline{\mathcal{A}_{x_{<i}}}$ be the elements of $\mathcal{V}$ not in $\mathcal{A}_{x_{<i}}$.
Then we can define the \textit{support-weighted total variation distance} as
\par\nobreak
\vspace{-10pt}
{\small
  \setlength{\abovedisplayskip}{6pt}
  \setlength{\belowdisplayskip}{\abovedisplayskip}
  \setlength{\abovedisplayshortskip}{0pt}
  \setlength{\belowdisplayshortskip}{4pt}
\begin{align}
\text{TV}_S(P^*(X_i\mid x_{<i}), &P_\text{trunc}(X_i\mid x_{<i})) \\
= &\beta_{\text{var}} \sum_{x\in S_{x_{<i}}^*\cap \bar{\mathcal{A}}} P^*(x\mid x_{<i}) \nonumber \\
+ &\beta_{\text{sup}}\sum_{x\in \overline{S_{x_{<i}}^*}\cap \mathcal{A}} P_\text{trunc}(x\mid x_{<i}) \nonumber
\end{align}
}%
The first term represents the total probability mass of the true distribution lost to truncation, weighted by hyperparameter $\beta_{\text{var}}$.
The second term represents the total probability mass placed off the support of the true distribution (thus constituting a bad continuation), weighted by $\beta_{\text{sup}}$.\footnote{See Section~\ref{appendix:stv} for the relationship to the total variation distance.}

Since the mass of a word under the true model, $P^*(x\mid x_{<i})$, may be arbitrarily close to zero, it is hard to guarantee that the first term ($\beta_\text{var}$) is zero.
One cannot guarantee that any non-complete allowed set $\mathcal{A}$ contains the full support of $P^*$.
However, the smoothing model does provide bounds on the probabilities of words in $\overline{S_{x_{<i}}^*}\cap \mathcal{A}$, meaning we can in principle avoid unnecessarily truncating words while still maintaining zero cost from the $\beta_{\text{sup}}$ precision term.
While we cannot know the exact properties of the unobserved smoothing distribution, we can use this fact to design principles desmoothing algorithms should follow.

\subsection{Principles for truncation as desmoothing}
Our LM framing specifies bounds on the probabilities of words outside the support of the true distribution, and our TV$_S$ motivates minimizing the difference between the allowed set $\mathcal{A}_{x_{<i}}$ and the support $S^*_{x_{<i}}$.
We now use both of these to describe principles for truncation; if these principles are not met, the word is in the support of $S^*_{x_{<i}}$ and should not be truncated.

\paragraph{Absolute probability.}
Under our smoothing model (Section~\ref{section:smoothingmodel}), a word outside the support of $P^*(X_i\mid x_{<i})$ has a bound on its probability:
\par\nobreak
\vspace{-15pt}
{
  \setlength{\abovedisplayskip}{6pt}
  \setlength{\belowdisplayskip}{\abovedisplayskip}
  \setlength{\abovedisplayshortskip}{0pt}
  \setlength{\belowdisplayshortskip}{4pt}
\begin{align}
\max_{x\not\in S_{x_{<i}}^*} P_\theta(x\mid x_{<i}) \ \ \leq (1+\delta)(1-\bar{\lambda})/|\mathcal{V}|,
\end{align}
}%
since we posited that smoothing never accounts for more than $\bar{\lambda}$ of the distribution.
While these terms are not known, the bound is likely small (since $\delta$ is small).
Hence as a general principle, words with large probability should not be truncated, since above a small probability threshold, they must be in the support of $P^*$.

\paragraph{Relative probability.}
Under our model, a distribution with high entropy has less smoothing, that is, $\lambda$ is smaller, e.g., note the term $\exp(-h_{x_{<i}})$ in the bound on $\lambda$.
This directly results in a lower maximum probability a word outside the support of the true distribution can achieve:
{
  \setlength{\abovedisplayskip}{6pt}
  \setlength{\belowdisplayskip}{\abovedisplayskip}
  \setlength{\abovedisplayshortskip}{0pt}
  \setlength{\belowdisplayshortskip}{4pt}
\begin{align}
\max_{x\not\in S_{x_{<i}}^*} P_\theta(x) \ \ \leq \alpha\exp(-h),
\end{align}
}%
where $\exp(-h_{x_{<i}})$ is the probability of a word in the uniform distribution of entropy $h_{x_{<i}}$ (and $\alpha$ is a constant).
The general principle is to only truncate words whose probabilities are also low relative to the rest of the distribution.

\subsection{Desmoothing and $n$-gram models}
The issue of smoothing on sample quality is apparent in $n$-gram language models.
An $n$-gram language model MLE estimate explicitly counts the number of times each $(n-1)$-word phrase is followed by a word in $\mathcal{V}$.
To avoid infinite perplexity (as the count estimates are zero almost everwhere), an $n$-gram model is explicitly smoothed \cite{katz1987estimation,church1991comparison}.

Text generated from \textit{unsmoothed} $n$-gram models is locally coherent.\footnote{As noted by Yoav Goldberg \url{https://nbviewer.org/gist/yoavg/d76121dfde2618422139} and \citet{jurafsky2000speech}, Chapter 3: N-gram Language Models.} 
However, we show that $n$-gram models smoothed with the uniform distribution generate nonsense (Figure~\ref{figure_ngram_samples}).
Why is this? Consider a $5$-gram LM smoothed with the uniform distribution.
If $x'$ is sampled from outside the support of the 5-gram model's support, then the new history $(x_{i-1}, x')$ \textit{was never seen during the training of the 5-gram model}, so now the model has only the poorly estimated probabilities from the smoothing distribution.%

\section{Methods}
We now describe in detail two popular truncation sampling algorithms, discuss how they break our desmoothing principles, and then present two new truncation sampling algorithms including our proposed $\eta$-sampling.

\subsection{Top-$p$ (nucleus) sampling}
Top-$p$ (nucleus) sampling truncates words that are outside the mimimal set of (most probable) words that account for at least $p$ percent of the distribution.
That is, the allowed set is as follows.
Let $x^{(1)}, \dots, x^{(|\mathcal{V}|)}$ be the words in $\mathcal{V}$ sorted in order of decreasing probability under $P_\theta(X \mid x_{<i})$.
Then let $j$ be the integer such that $j=\arg\min_{j'} \sum_{i=1}^{j'} P_\theta(x^{(i)}\mid x_{<i}) \geq p$.
The allowed set of top-$p$ sampling is then $\mathcal{A}_{x_{<i}} = \{x^{(1)},\dots, x^{(j)}\}$.\footnote{Often, $p$ is taken as 0.9 or 0.95.}
Top-$p$ sampling breaks the absolute probability principle: words with up to $(1-p)$ probability may be truncated simply because other high-probability words cover probability $p$.
For the prompt \textit{My name}, the word \textit{is} is assigned $0.96$ probability by GPT-2, but less likely candidates \textit{'s}, \textit{was} and \textit{isn} shouldn't be truncated.
Intuitively, $(1-p)$, e.g., $0.05$ or $0.01$ is quite high probability given a vocabulary size of, e.g., 50,000.

\subsection{Typical decoding}
Typical decoding is motivated by local informativeness: never generate words that are too surprising or too predictable \cite{meister2022typical}.
The algorithm sorts the vocabulary in order of the difference between the entropy $h_{\theta,x_{<i}}$ of the LM conditional distribution and the negative log-probability of the word, and takes words from this list to cover $p$ percent of the distribution.
That is, let $x^{(1)}, \dots, x^{(|\mathcal{V}|)}$ be the words in $\mathcal{V}$ in sorted order of increasing $|h_{\theta,x_{<i}}+\log p_\theta(x\mid x_{<i})|$.\footnote{$h_{\theta,x_{<i}}=-\sum_{x\in\mathcal{V}} P_\theta(x\mid x_{<i})\log P_\theta(x\mid x_{<i} )$.}
Then let $j$ be the integer $j=\arg\min_{j'} \sum_{i=1}^{j'} P_\theta(x^{(i)}\mid x_{<i}) \geq p$.
The allowed set of typical decoding is $\mathcal{A}_{x_{<i}} = \{x^{(1)},\dots, x^{(j)}\}$.
This breaks the absolute probability principle for the same reason as top-$p$, and additionally can truncate the most probable words.

\subsection{$\epsilon$-sampling (ours)}
The absolute probability principle---that words outside the support of the true distribution have low probability---suggests a simple truncation algorithm: for some hyperparameter threshold $\epsilon$ allow any word with greater than $\epsilon$ probability.
\begin{align}
\mathcal{A}_{x_{<i}} = \{ x\in\mathcal{V} : P_\theta(x \mid x_{<i}) > \epsilon\}
\end{align}
In the case of the prompt \textit{My name} where top-$p$ rejects plausible words because of the probability assigned to \textit{is} (and \textit{'s}), $\epsilon$-sampling allows additional words with a threshold of, e.g., $0.0003$.

However, $\epsilon$-sampling breaks the relative probability principle.
For example, the prompt \textit{The} should allow many continuations, and top-$p$ with GPT-2 allows over ten thousand words, but $\epsilon$ would have to be impractically small to do so.
This is a key failure akin to that of top-$k$ sampling; when many next words are plausible, the allowed set should reflect that.

\subsection{$\eta$-sampling (ours)}

Our proposed algorithm, $\eta$-sampling, composes respect for both the absolute and relative probability principles.
Consider a conditional distribution $P_\theta(X\mid x_{<i})$ with entropy $h_{\theta,x_{<i}}$.
The probability of a word in the uniform distribution of entropy $h_{\theta,x_{<i}}$ is $\exp(-h_{\theta,x_{<i}})$.
Our entropy-dependent threshold is $\alpha\exp(-h_{\theta,x_{<i}})$ where $\alpha\in[0,1]$.
Combining this rule with our epsilon rule for the absolute probability principle, we come to:
\begin{align*}
&\mathcal{A}_{x_{<i}} = \{ x \in \mathcal{V} \mid P_\theta(x \mid x_{<i}) > \eta \}\\
&\eta = \min \big(\epsilon, \alpha\exp(-h_{\theta,x_{<i}})\big)\}
\end{align*}
where $h_{\theta,x_{<i}}$ is the entropy of $P_\theta(X \mid x_{<i})$.
In this work, to expose a single hyperparameter, we set $\alpha=\sqrt{\epsilon}$, which we find works well empirically. For intuition, think of $\epsilon\approx0.0009$. 

\paragraph{Analysis of $\eta$-sampling.} Returning to our smoothing model, we note that $\eta$-sampling approximates optimal desmoothing in the regime that the support penalty $\beta_{\text{sup}}$ dominates the variation penalty $\beta_{\text{var}}$.
Consider a truncation algorithm that truncates as $\eta$-sampling, but sets $\eta$ as:
\par\nobreak
\vspace{-10pt}
{\small
  \setlength{\abovedisplayskip}{6pt}
  \setlength{\belowdisplayskip}{\abovedisplayskip}
  \setlength{\abovedisplayshortskip}{0pt}
  \setlength{\belowdisplayshortskip}{4pt}
\begin{align}
&\eta = \min \big(\frac{(1-\bar{\lambda})(1+\delta)}{V}, \alpha\exp(-h_{x_{<i}})\big)\},
\end{align}}
where $h_{x_{<i}}$ is the entropy of the true distribution, not $P_\theta$.
We're guaranteed that the support loss (the term weighted by $\beta_{\text{sup}}$) is zero, and that the variation loss (weighted by $\beta_{\text{var}}$) is minimized relative to the constraint of zero support loss.
 If $x\not\in S_{x_{<i}}^*$, then the probability of $x$ is less than or equal to the min of $(1-\bar{\lambda})(1+\delta)/V$ and $\frac{V\alpha\exp(-h_{x_{<i}})}{1+\delta} \times \frac{1+\delta}{V}=\alpha\exp(-h_{x_{<i}})$.
So, we're guaranteed that $\mathcal{A}_{x_{<i}}\subseteq S_{x_{<i}}^*$, and truncating more would break this guarantee.\footnote{See Appendix~\ref{appendix_analysis} for an expanded version of this argument.}
Our $\eta$-sampling approximates this by using the LM entropy instead of the unavailable true distribution entropy, and without knowing the true hyperparameters.

\section{Experiments \& Results}
Our experiments characterize $\eta$-sampling relative to the state-of-the-art top-$p$ and typical decoding.
We use MAUVE, an automatic metric for open-ended generation, to find hyperparameters giving comparable diversity-accuracy tradeoffs.
$\eta$-sampling behaves better in a range of settings, from long-document generation to more defensibly truncating low-entropy distributions.

\paragraph{Models \& Data.}
In all experiments, we use all or some subset of the four GPT-2 models \cite{radford2019language} of varying sizes.
Experiments are run on in-distribution, held-out data from the validation or test set of GPT-2 (WebText), since it is composed of a wide variety of long-form documents.

\subsection{Hyperparameter sweep on MAUVE}
\label{sec_hyperparams}

\begin{table}
\centering
\small
\begin{tabular}{c l}
\toprule
\bf Method & \bf Hyperparameters \\
\midrule
top-$p$ & \{0.89, 0.9, 0.92, 0.95, 0.99\}\\
typical & \{0.2, 0.9, 0.92, 0.95\} \\
$\epsilon$ & \{0.001, 0.0009, 0.0006, 0.0003, 0.0001\} \\
$\eta$ & \{0.004, 0.002, 0.0009, 0.0006, 0.0003\}\\
\bottomrule
\end{tabular}
\caption{\label{table_hyperparams} Hyperparameter sweep for each method.}
\end{table}

\begin{table}
\centering
\small
\begin{tabular} {@{} l r r r r @{}}
\toprule
\textbf{Method\ \ \textbackslash\ \ Model} & \textbf{sm} & \textbf{med} & \textbf{lg} & \textbf{xl}\\
\midrule
raw sampling $\dagger$  & 0.589 & 0.373 & 0.845 & 0.882 \\
top-$p$ $\dagger$ & 0.878 & 0.915 & \bf 0.936 & 0.940 \\
top-$p$ (our replication) & 0.874 & 0.917 & 0.932 &\bf 0.944 \\
Typical Decoding & 0.873 & 0.906 & 0.922 & 0.939 \\
\midrule
$\epsilon$-sampling (ours) &0.874 &0.918 & \bf 0.936 & 0.941  \\
$\eta$-sampling (ours) & \bf 0.880 &  \bf 0.920 & 0.935 &  0.942 \\
\bottomrule
\end{tabular}
\vspace{4pt}
\caption{\label{table_mauve} Results on the MAUVE metric for open-eneded GPT-2 WebText generation. Higher is better. The $\dagger$ indicates numbers drawn from \citet{pillutla2021mauve}. Bold indicates best for model, not necessarily significantly.}
\end{table}

We first find hyperparameters for each of top-$p$, typical decoding, $\epsilon$-sampling, and $\eta$-sampling that maximize MAUVE score for each GPT-2 model on WebText.

\paragraph{Setting.}
Following the MAUVE paper's setting exactly \cite{pillutla2021mauve}, we take the GPT-2 family of models and 5,000 samples from their test data. For each sample, we prompt the model with 35 words and generate until at most 1024 words.
We study GPT-2 small (124M parameters), medium (355M), large (774M) and XL (1.5B) models.

\paragraph{Evaluation.}
MAUVE attempts to measure both the precision (are samples generally like those from the true distribution) and recall (is the variability in samples like that of those from the true distribution) of samples from a text generation system.
It was shown by \citet{pillutla2021mauve} to correlate well with human judgments.

\paragraph{Hyperparameters.}
Top-$p$, typical decoding, $\epsilon$-sampling, and $\eta$-sampling all have a hyperparemter which determines the severity of truncation.
The set we search over is given in Table~\ref{table_hyperparams}.\footnote{The hyperparameter set for our methods was chosen to have similar average total variation values between pre- and post-truncation to the top-$p$ set.}
We pick the best hyperparameter using 2--5 seeds on the validation set, and report the average performance across 5 seeds on the test set.

\paragraph{Results.}
The results are reported in Table~\ref{table_mauve}; we find that overall, the methods perform similarly, with typical decoding performing slightly worse than top-$p$ and our methods.

\begin{figure*}
\centering
\includegraphics[width=0.32\linewidth]{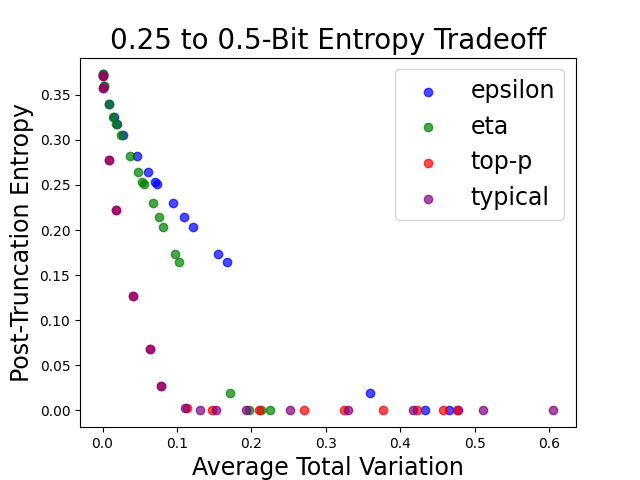}
\includegraphics[width=0.32\linewidth]{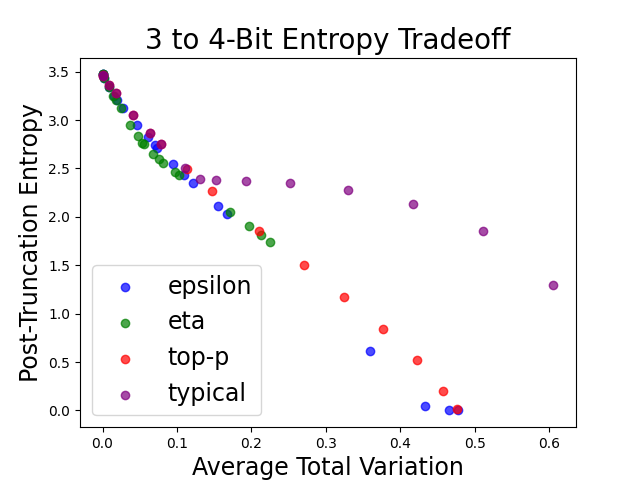}
\includegraphics[width=0.32\linewidth]{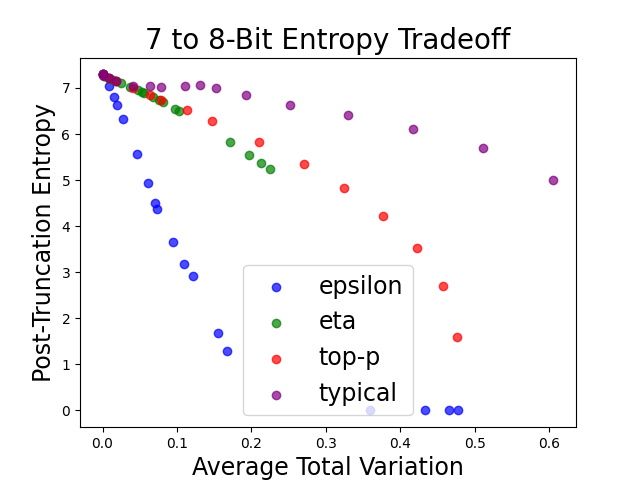}
\caption{\label{figure_entropy_plots}Top-$p$ sampling aggressively truncates low-entropy distributions and $\epsilon$-sampling aggressively truncates high-entropy distributions, while $\eta$-sampling strikes a balance.}
\end{figure*}

\subsection{Human evaluation of long-document suffix plausibility}

\begin{table}
\centering
\small
\begin{tabular}{@{} l c c c @{}}
\toprule
& \multicolumn{3}{c}{\it Study 1: Human vs top-$p$ vs $\eta$} \\
& top-$p$ & $\eta$-sampling & Human\\
\midrule
Top-$p$ vs human & 43 (43\%) & --- & \bf 56 (56\%) \\
$\eta$ vs human & --- & 42 (42\%) & \bf 53 (53\%) \\
Top-$p$ vs $\eta$ & 39 (39\%) &\bf 53 (53\%) & ---  \\[1ex]
\toprule
& \multicolumn{3}{c}{\it Study 2: top-$p$ vs $\eta$-sampling} \\
& Top-$p$ & $\eta$-sampling & Equal \\
\midrule
Top-$p$ vs $\eta$ & 118 (40\%) & \bf 159 (53\%) & 17 (6\%) \\
\bottomrule
\end{tabular}
\caption{\label{table_humaneval}Human preferences of long-document plausibility; we report absolute numbers of judgments, and percentages in parentheses. Judgment percents that both suffixes were too bad to judge can be inferred.}
\end{table}

We now study whether $\eta$-sampling leads to more coherent long-document generations than top-$p$ sampling.
We omit typical decoding since it does not seem to outperform top-$p$ on MAUVE\@.
Considering that holistic evaluation of long texts is difficult for humans \cite{ippolito-etal-2020-automatic} we design a human study to evaluate \textit{long document plausibility}: given a shared document prefix, which method's generated suffix (omitting the middle) is more reasonably from the same document?
This new evaluation avoids forcing humans to keep up to 1024 words in working memory.%

\paragraph{Setting.} 
For each of top-$p$ and $\eta$-sampling, we sample from GPT-2 large with MAUVE-maximizing hyperparameters, conditioned on each prefix of 35 subword tokens from the WebText validation set.
From this set we filter to prefixes for which the reference and both generated documents are at least 900 tokens long and pass manual filter for quality.\footnote{We also manually filter prompts for quality, following \citet{pillutla2021mauve}. See Appendix~\ref{appendix_filter}.}
59 workers from the United States were recruited on Amazon Mechanical Turk with the Master qualification, and paid $\$1$ per task with an expected time of 3.5 to 4 minutes.
We run two studies.

\paragraph{Study 1.}
We show a human evaluator the 35-token prefix, as well as the last 70 tokens of two documents (of the 3 possible).
The evaluator is asked to judge which of the two suffixes may \textit{more reasonably be from the same document} as the prefix, or to note that both are too bad to judge.
For each of the three possible pairings of top-$p$, $\eta$-sampling, and reference document, we elicit 100 human judgments over 100 prefixes. 

\paragraph{Study 2.}
We ran a second study just comparing top-$p$ to $\eta$-sampling to allow for larger $n$, since we had finite resources and the result that both methods generate text worse than humans is not at issue.
To test whether the effect size observed was in part due to forcing evaluators to pick one of the two methods, in this study we allow human evaluators to mark that both suffixes are of equal quality.

\paragraph{Results.}
The results are reported in Table~\ref{table_humaneval}.
In Study 1, we find that human document generations are preferred over top-$p$ and $\eta$-sampling at roughly the same rate, while $\eta$-sampling is preferred over top-$p$ (53\% to 40\%).
In Study 2, we find that $\eta$-sampling is significantly preferred more frequently than top-$p$ with a Wilcoxon paired test ($p=0.0138$) at the same effect size.

\subsection{Entropy analysis}
We now want to build a deeper understanding of the characteristics of the algorithms: what parts of the distribution tend to get cut by each method?
In our first analysis, we study whether each method has a tendency to aggressively truncate distributions of a given entropy.
A low-entropy distribution might be given by the prompt \textit{Barack Obama went to the White \ldots}, while a high-entropy distribution might be given by the prompt \textit{My name is \ldots}.

\paragraph{Setting.}
For a range of hyperparameters, we plot the average amount of truncation across all contexts against the retained entropy for an entropy range.
We use total variation to measure average truncation, $\mathbb{E}_{x_{<i} \sim P(X)} \|P_\theta(X_i\mid x_{<i}) - P_\text{trunc}(X_i\mid x_{<i})\|_{\text{TV}}$.
For each entropy range $R$, we consider the set $\mathcal{X}_R$ of prefixes $x_{<i}$ with pre-truncation entropy $h_{\theta,x_{<i}}$ in $R$ and compute the average remaining entropy $\frac{1}{|\mathcal{X}_R|}\sum_{\mathcal{X}_R} h_{\text{trunc},x_{<i}}$ after truncation.

\paragraph{Results.} The results for GPT-2 XL are presented in Figure~\ref{figure_entropy_plots}.
We find that top-$p$ sampling heavily truncates low-entropy distributions compared to $\epsilon$-sampling and $\eta$-sampling.
$\epsilon$-sampling heavily truncates high-entropy distributions. 
Typical behaves like top-$p$ for low-entropy distributions, and retains more entropy in high-entropy distributions.\footnote{This is likely because typical decoding cuts the non-uniform head of the distribution, and keeps the more-uniform middle.}
$\eta$-sampling strikes a good balance of not heavily truncating low- or high-entropy distributions.

\begin{table}
\small
\centering
\begin{tabular}{l c c c c}
\toprule
& \multicolumn{4}{c}{\bf Repetition Percent} \\
\textbf{Truncation \ \textbackslash \ Model} & \textbf{sm} & \textbf{med} & \textbf{lg} & \textbf{xl}\\
\midrule
top-$p$ & 54\% & 61\% & 47\%& 27\% \\
typical & 51\% & 61\% & 56\% & 37\%\\
$\epsilon$-sampling (ours) & \bf 28\% & \bf 37\% &   \bf 23\% & \bf 11\% \\
$\eta$-sampling (ours) &  37\% & 40\%& 26\%  &  12\%\\
\bottomrule
\end{tabular}
\caption{\label{table_repetition}Table showing repetition-degeneration rates for each method in an adversarial setting; lower is better.
}
\end{table}

\begin{figure*}
\centering
\includegraphics[width=\linewidth]{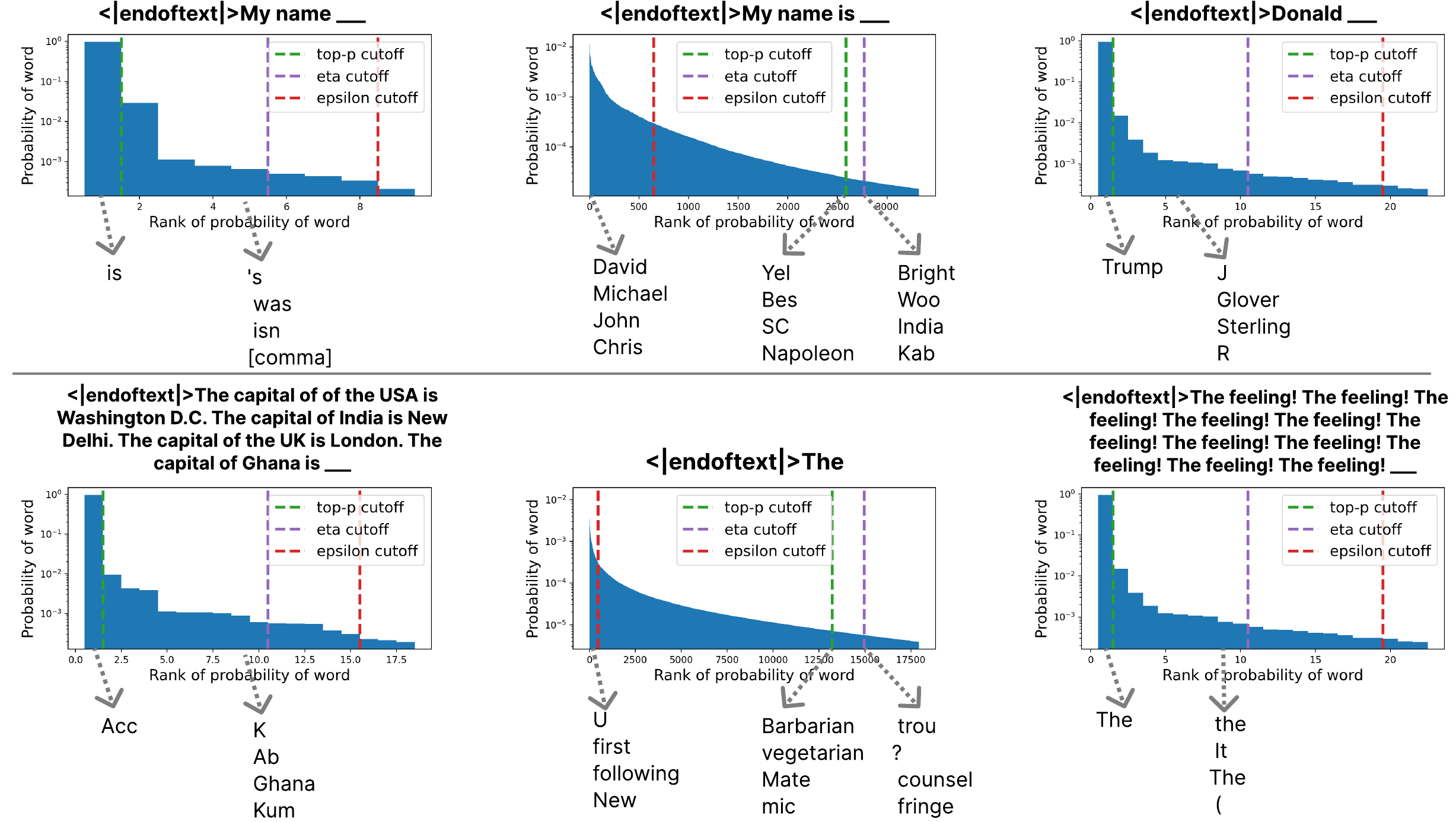}
\caption{\label{figure_names}Unit tests of the truncation behavior of top-$p$, $\epsilon$, and $\eta$-sampling on CheckList-inspired prefixes.}
\end{figure*}

\subsection{Repetition analysis}

We hypothesize that the tendency of top-$p$ sampling to heavily truncate low-entropy distributions causes it to generate repetitive text by only allowing the repetition-continuing word. %
To stress test the methods, we devise an adversarial setting in which the prompt has repetitions (as may be the case due to noisy input or natural repetition) and then determine whether the methods break the repetition.

\paragraph{Setting.} We take natural prompts---the first 35 words of the Wikipedia biographies of the 101 people with the most-read Wikipedia pages---and synthetically corrupt them by repeating the last 3 subword tokens 5 additional times.
Even with the existing repetition in the prompt, we want models to break the cycle and generate normal text again.
Here's an example prompt:
\begin{quote}
\small
Shawn Corey Carter (born December 4, 1969), known professionally as Jay-Z, is an American rapper, songwriter, record executive, entrepreneur, and media proprietor and media proprietor and media proprietor and media proprietor and media proprietor and media proprietor
\end{quote}
For each prompt, we generate 5 completions of up to 512 words.
For each of the GPT-2 models, we take the hyperparameter for each truncation sampling algorithm from Section~\ref{sec_hyperparams}, and compute the percent of completions that continue to repeat.\footnote{
Any sample with less than $1$ average negative log probability under the model is labeled a repetition.
We found this more useful than $n$-gram repetition statistics, as, e.g., repetition can involve small variation.}

\paragraph{Results.}
$\epsilon$-sampling achieves the lowest repetition rate, with e.g., 23\% for GPT-2 large, while $\eta$-sampling performs slightly worse (e.g., 26\%).
Top-$p$ causes considerably more repetition (e.g., 47\%).
Typical sampling causes slightly more repetition than top-$p$.\footnote{This is likely because the MAUVE-maximizing hyperparameter for typical sampling (e.g., $0.92$ for GPT-2 large) is generally more conservative than that for top-$p$ (e.g., $0.95$.)}

\subsection{Studying individual distributions}

We now study specific truncation decisions made by each algorithm, to provide more detailed behavioral insights.
We construct prompts and observe the truncation behavior of each algorithm on the resulting distribution, treating each as a CheckList-like unit test \cite{ribeiro-etal-2020-beyond}.

\paragraph{Setting.} We take the GPT-2 large model, provide it with each of 6 prompts, and using the MAUVE-maximizing hyperparameters we found in Section~\ref{sec_hyperparams}, truncate the resulting distribution.
The prompts are shown in Figure~\ref{figure_names}.
For this experiment we only study top-$p$, $\epsilon$, and $\eta$-sampling.

\paragraph{Results.}
The results are visualized in Figure~\ref{figure_names}.
We use two low-entropy prompts, \textit{My name...} and \textit{Donald...} and in both cases, find that top-$p$ decoding only allows a single word continuation. 
Top-$p$ can only generate \textit{is} after \textit{My name}, and \textit{Trump} after \textit{Donald}, which we find undesirable; we would like our truncation to allow, e.g., multiple Donalds to be discussed.
For a prompt with the phrase \textit{The feeling!} repeated multiple times (as one might say euphorically), top-$p$ can only continue the repetitive pattern, unlike $\epsilon$ and $\eta$-sampling.
For a prompt suggesting specification of capitals of countries, we find that top-$p$ only allows the \textit{correct} capital name, whereas $\eta$-sampling and $\epsilon$-sampling allow different continuations which do not follow the in-context trend, suggesting that top-$p$ may be better for generating, e.g., answers to questions.
We use two high-entropy prompts, \textit{The...} and \textit{My name is...}, finding that $\eta$-sampling and top-$p$ sampling allow a range of possibilities, unlike $\epsilon$-sampling.
The behavior of $\epsilon$-sampling in allowing \textit{fewer} words in higher entropy conditional distributions is a clear failure.

\section{Related Work}

\paragraph{Stochastic decoding algorithms.}
Stochastic decoding algorithms produce sequences from a model and involve randomness.
The simplest is \textit{sampling}, sometimes called \textit{ancestral sampling}, \cite{bishop2006pattern}, which generates a sample from the model.
Some stochastic decoding methods attempt to find high-likelihood sequences instead of attempting to recreate the true distribution, like stochastic beam search \cite{kool2019stochastic} and conditional poisson stochastic beam search \cite{meister-etal-2021-conditional}.
Truncation sampling algorithms, like top-$k$ \cite{fan-etal-2018-hierarchical}, top-$p$ \cite{Holtzman2020The}, and Mirostat \cite{basu2021mirostat}, are intended to improve quality but keep variety.
\citet{welleck2020consistency} found that truncation algorithms can lead to non-zero mass assigned to infinite sequences.

\paragraph{KL-divergence, language models, smoothing.}
The most famous example of methods that do not cover every mode is GANs \cite{goodfellow2014generative}.
In language modeling, some have pointed to the inability of the softmax function to assign 0 probability to any category as a deficiency and proposed sparse alternatives \cite{martins2016softmax,peters2019sparse,tezekbayev2021speeding}.
This intuition is akin to ours, as is loss truncation \cite{kang-hashimoto-2020-improved}, which keeps rare events from incurring arbitrarily high loss.
\citet{mohri-roark-2006-probabilistic} attempt to identify structural zeros in the distribution of language when inducing probabilistic context-free grammars.

\paragraph{High-entropy language generation \& evaluation.}
Evaluation of open-ended generation of natural language is difficult; one must evaluate both the quality of samples and the diversity.
Quality is hard to measure in high-entropy generation, and is often not correlated with model probability \cite{hashimoto2019unifying,meister-etal-2022-high}.
An emergent line of work connects human notions of quality, and human generative tendencies, with the uniform information density hypothesis (e.g., leading to typical decoding) \cite{wei-etal-2021-cognitive,meister-etal-2021-revisiting}.
Both \citet{meister-cotterell-2021-language} and \citet{pillutla2021mauve} directly estimate whether model samples' statistics match those of natural language.
\citet{nadeem2020systematic} study properties held by successful strategies for reallocating mass away from the tail of LM distributions.

\section{Conclusion}

We've framed the class of truncation sampling algorithms as performing desmoothing, an insight that led to principles for how truncation should be done to recover the training distribution, a new truncation sampling algorithm, and evaluations that show the deficiencies of existing algorithms.
We find the tendency of top-$p$ decoding to over-truncate low-entropy distributions to be particularly surprising.
We aim for these insights, and the evaluations we use, to drive further research in understanding and improving how we generate from neural language models.

\section*{Acknowledgements}
The authors would like to thank John Thickstun, Rishi Bommasani, Kaitlyn Zhou, Will Merrill, Nelson Liu, and Tatsunori Hashimoto for helpful discussions on this work, and to the reviewers for clarifying feedback.
JH was supported by an
NSF Graduate Research Fellowship under grant
number DGE-1656518.
We gratefully acknowledge the support of a PECASE Award.

\section{Limitations}
With the analysis we've done, we believe it to be very difficult to derive an understanding of all the \textit{sequence-level} effects truncation sampling algorithms (including ours) have: what kinds of sequences are we disallowing?
What types, or sources of language are being (unknowingly) disallowed?
Beyond this, we've only tested our algorithms on English language models; the conditional distributions of languages with rich morphology likely have different properties (especially with subword models).

\section{Ethics Statement}
Any work to improve generative models of text comes with ethical concerns surrounding negative use cases of text generation including hate speech and misinformation.
While our algorithm does improve long text generation, we hope it also provides insight into the unintended and until-now unknown consequences of existing truncation sampling algorithms (including top-$p$).
Algorithms like ours, which reallocate probability mass from the least likely elements of a distribution, have a particular risk of harm in removing the ability of models to talk about topics or names that are already rare.
Concurrent work finds that the choice of stochastic decoding algorithm affects measured fairness metrics in open-ended generation \cite{dhamala2018analysis}.
Our framing, and the hope for future work, is to use truncation to recover something as close to the \textit{training} distribution as possible; of course, the training distribution must then be chosen with care. Generating a word due to smoothing (noise) would likely mean that subsequently generated words about that topic would be low-quality, which is also undesirable.

\bibliography{anthology,custom}
\bibliographystyle{acl_natbib}

\appendix

\section{Notes}

\subsection{Support-weighted total variation}
\label{appendix:stv}
We introduce new notation just for this section, to present support-weighted total variation in generality.
Recall that the total variation distance between  discrete distribution $R$ over space $\mathcal{V}$ and discrete distribution $U_t$, the result of truncation with allowed set $\mathcal{A}\subseteq \mathcal{V}$ from a discrete distribution $U$ over $\mathcal{V}$ , is
\begin{align}
\sum_{x\in\mathcal{V}} | R(x) - U_t(x) |.
\end{align}
Denoting the support of $R$ as $S_R$, we can partition $\mathcal{V}$ into four sets:
\begin{align}
&S_R \cap \mathcal{\bar{A}} \nonumber \\
&\overline{S_R} \cap \mathcal{A} \nonumber \\ 
&S_R \cap \mathcal{A} \nonumber \\
&\overline{S_R} \cap \bar{\mathcal{A}} 
\end{align}
We split the sum of the total variation distance into these four terms.

The first represents the words that are in the support of $R$ but not in the allowed set of $U_t$:
\begin{align}
\sum_{S_R \cap \mathcal{\bar{A}}} | R(x) - U_t(x) | = \sum_{S_R \cap \mathcal{\bar{A}}} R(x),
\end{align}
since $U_t(x)=0$ if $x\not\in\mathcal{X}$.
This exactly represents the total probability mass that was lost from $R$.
The second term represents the words that are not in the support of $R$ but were allowed:
\begin{align}
\sum_{\overline{S_R} \cap \mathcal{A}} | R(x) - U_t(x) | = \sum_{\overline{S_R} \cap \mathcal{A}} U_t(x),
\end{align}
since $R(x)=0$ if $x\not\in S_R$.
This exactly represents the total probability that we sample a word from $U_t$ that has zero probability under $R$ (and so we move off the support of $R$ for future generation.)
the third term is the words that were correctly allowed:
\begin{align}
\sum_{S_R \cap \mathcal{A}} | R(x) - U_t(x) |.
\end{align}
In this case, $U_t(x)$ may be an under or overestimate of $R(x)$.
The last term is the words that were correctly truncated:
\begin{align}
\sum_{\overline{S_R} \cap \bar{\mathcal{A}}} | R(x) - U_t(x) | = \sum_{\overline{S_R} \cap \bar{\mathcal{A}}} | 0 - 0 |
\end{align}
which is identically zero.

To form our support-weighted total variation metric, we took the first two terms, which are interpretable and each exactly specifies one of the two desiderata from a truncation algorithm: maintaining the variety of $R$, and not generating a word that $R$ wouldn't generate.
However, in different use cases, one or the other may be more crucial; hence we give each its own hyperparameter, $\beta_{\text{var}}$ and $\beta_{\text{sup}}$, to arrive at our metric,
\begin{align}
\text{TV}_S(R, U_t) = &\beta_{\text{var}} \sum_{x\in S_R\cap \bar{\mathcal{A}}} R(x) \nonumber \\
+ &\beta_{\text{sup}}\sum_{x\in \overline{S_R}\cap \mathcal{A}} U_t(x).
\end{align}

\subsection{Analysis of $\eta$-sampling} \label{appendix_analysis}
The purpose of this analysis is to show that if one assumes our smoothing model, then an $\eta$-sampling approximates an algorithm that avoids sampling from outside the support of the true distribution while minimilly truncating the distribution.

Consider a conditional distribution from a language model under our model, $P_\theta(X_i\mid x_{<i})$.
Consider an allowed set $\mathcal{A}_{x_{<i}}$ defined via a probability threshold, $\mathcal{A}=\{x \mid P_\theta(x\mid x_{<i}) > \eta^*\}$, where $\eta^*$ is defined as
\par\nobreak
\vspace{-10pt}
{\small
  \setlength{\abovedisplayskip}{6pt}
  \setlength{\belowdisplayskip}{\abovedisplayskip}
  \setlength{\abovedisplayshortskip}{0pt}
  \setlength{\belowdisplayshortskip}{4pt}
\begin{align}
&\eta^* = \min \big(\frac{(1-\bar{\lambda})(1+\delta)}{V}, \alpha\exp(-h_{x_{<i}})\big)\}.
\end{align}}
In this case, it is guaranteed that $x\in S_{x_{<i}}^*$, since $\eta^*$ represents the maximum probability of a word whose probability stems entirely from the smoothing distribution.

If one sets a lower probability threshold $\eta' = \eta^* - \psi$ for some $\psi > 0$ when computing the allowed set, then under our model, there can be a conditional distribution such that $x\not\in S_{x_{<i}}^*$, and $P_\theta(x\mid x_{<i})>\eta'$.
Such an $x$ would be incorrectly allowed.

Similarly, if one sets a higher probability threshold $\eta' = \eta^* + \psi$ for some $\psi > 0$ when computing the allowed set, then under the model, there can be a conditional distribution such that $x\in S_{x_{<i}}^*$, and $P_\theta(x \mid x_{<i}) \in (\eta, \eta')$.
Defining the allowed set with $\eta'$, we truncate $x$, which is unnecessary, since words in $S_{x_{<i}}^*$ have probability at least $\eta$ under the language model.

This argument has considered truncation algorithms that specify their allowed set as every word in $\mathcal{V}$ with LM probability above a threshold, showing that setting the threshold as $\eta^*$ guarantees (under our model) that we sample from the support of the true distribution without unnecessarily truncating too much. 
We now consider allowed set defined by algorithms other than probability thresholds.
Let the allowed set defined according to the $\eta^*$ threshold be $\mathcal{A}^*_{x_{<i}}$.
Consider an allowed set $\mathcal{A}_{x_{<i}}$ defined by another truncation sampling algorithm (which may not define it via a probability threshold like.
If $\mathcal{A}_{x_{<i}}=\mathcal{A}^*_{x_{<i}}$, then the two algorithms are indistinguishable for this prefix.
Otherwise, if $x\in\mathcal{A}_{x_{<i}}$ and $x\not\in\mathcal{A}^*_{x_{<i}}$, then $x$ may be outside the support of the true distribution, and should have been truncated.
And if $x\in \mathcal{A}^*_{x_{<i}}$ and $x\not\in\mathcal{A}_{x_{<i}}$, then $x$ was unnecessarily truncated.

When using our $\eta$-sampling algorithm, we neither know the true hyperparameters, nor do we have access to the true distribution conditional entropy, so $\eta$-sampling only approximates this.
Specifically, we set the hyperparameters of $\eta$-sampling via search on the task of interest, and we use the observed LM entropy instead of the true distribution entropy in computing the relative probability threshold.
In practice, one wants to set a threshold of truncation based on the needs of the task and the tolerance for error, so a threshold that perfectly excludes words outside the true distribution support may not be optimal for the task of interest anyway.

\section{More Experimental Details}

\begin{table}
\small
\begin{tabular}{l r r r r}
\toprule
\bf Method \textbackslash Model & small & med & large & XL\\
\midrule
Top-$p$ & 0.9 & 0.89 & 0.95 & 0.95 \\
Typical & 0.9 & 0.9 & 0.92 & 0.92 \\
$\epsilon$-sampling & 0.0006 & 0.0009 & 0.0003 & 0.0003\\
$\eta$-sampling & 0.002 & 0.0006 & 0.0006 & 0.0003\\
\bottomrule
\end{tabular}
\caption{\label{table:hyperparam_choices} Best-performing hyperparameters according to MAUVE from experiments in Section~\ref{sec_hyperparams}.}
\end{table}

\subsection{Hyperparameters}
The MAUVE-maximizing hyperparameters for each truncation sampling algorithm for each model are provided in Table~\ref{table:hyperparam_choices}.

\subsection{$5$-gram model}
For our small demo demonstrating the behavior of smoothed $n$-gram models, we trained a $5$-gram model on 10,000 documents from The Pile \cite{gao2021pile}.
We smoothed the model with the uniform distribution.

\begin{figure*}
\includegraphics[width=\linewidth]{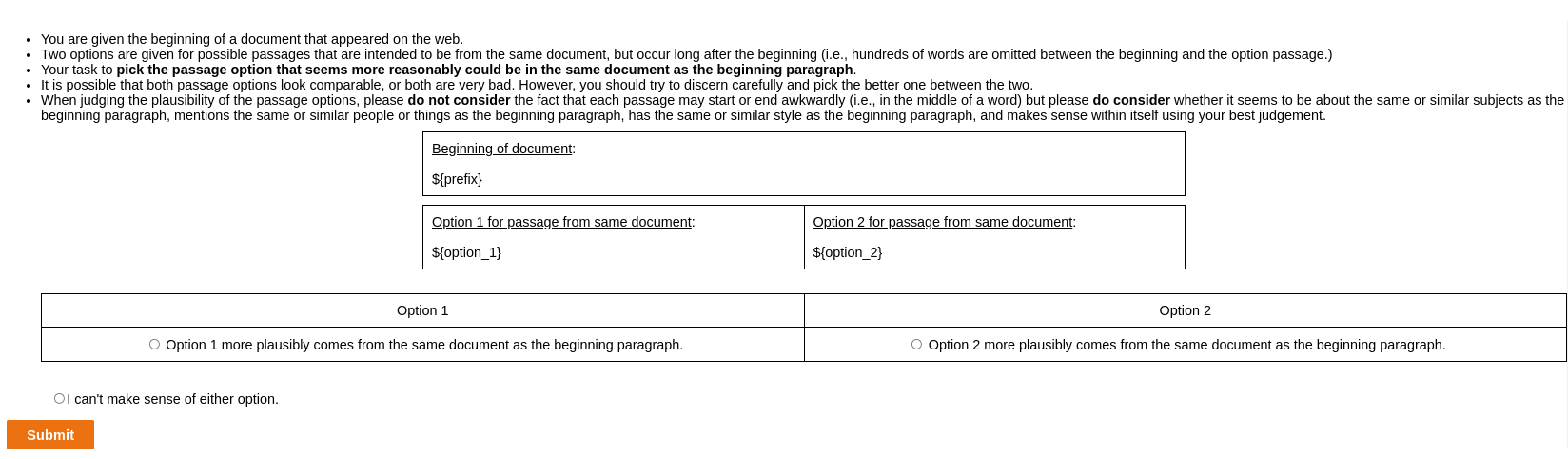}
\caption{\label{figure:study_amt1}The interface shown to human annotators for Study 1.}
\end{figure*}

\begin{figure*}
\includegraphics[width=\linewidth]{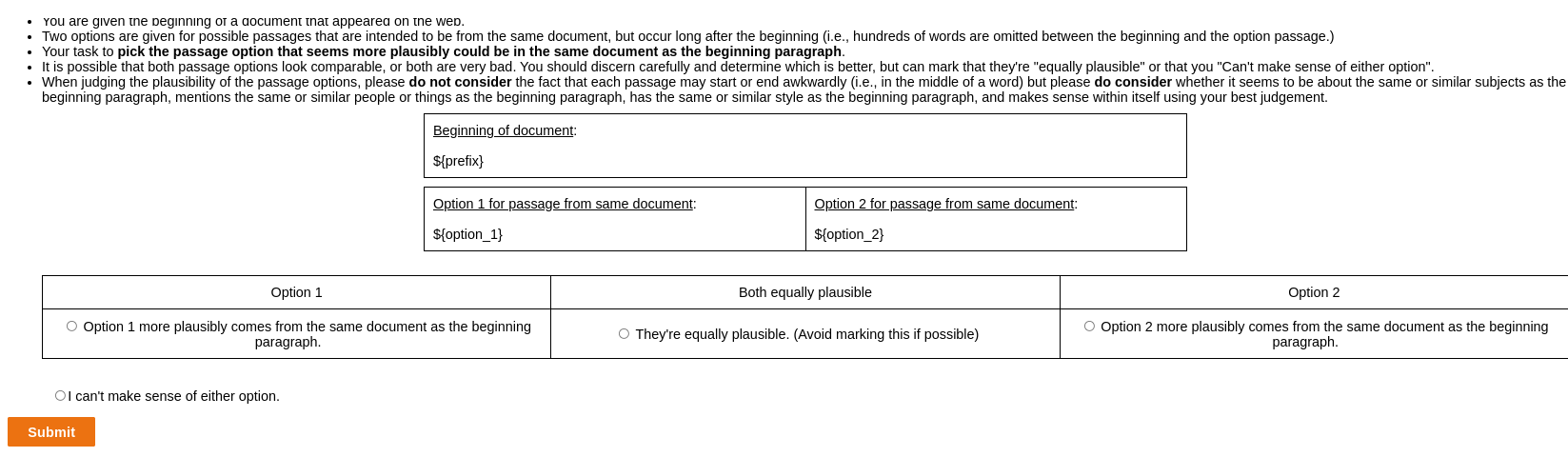}
\caption{\label{figure:study_amt2}The interface shown to human annotators for Study 2.}
\end{figure*}

\subsection{Amazon Mechanical Turk Details} \label{appendix_filter}
To provide more transparency into our human studies, we provide the form that was shown to human annotators for both of our studies.
The (similar) interfaces shown for Study 1 and Study 2 are shown in Figure~\ref{figure:study_amt1} and Figure~\ref{figure:study_amt2}, respectively.
We randomize the ordering of presentation of the methods' generations (note that the forms say ``Option 1'' and ``Option 2''.)

Of the 59 unique workers, 44 unique workers participated in study 1, and 36 unique workers participated in study 2.

We follow \citet{pillutla2021mauve} in manually filtering the WebText prompts that go into our human study.
Webtext is noisy, and not all prompts are clearly natural language.
Our manual filtering of prompts led to 36 rejected prompts (of 146 considered) due to quality for study 1.
Our manual filtering of prompts led to 100 rejected prompts (of 402 considered) due to quality for study 2.
This is compared to rejecting 3169 of 5000 prompts due to quality in the original MAUVE paper; we attempted to minimally filter while guaranteeing that prompts were natural language.
Our kept and filtered prompts are available in our codebase.

\end{document}